\documentclass[letterpaper, 10 pt, conference]{ieeeconf}  

\IEEEoverridecommandlockouts                              

\usepackage{graphicx, amsmath, amssymb} 
\usepackage[table]{xcolor}
\usepackage[british, american]{babel}
\usepackage{booktabs}
\usepackage{times}
\usepackage{epsfig}
\usepackage{tabulary}
\usepackage{soul}
\usepackage{xspace}\xspace
\usepackage{adjustbox}
\usepackage{array}
\makeatletter
\let\NAT@parse\undefined
\makeatother
\usepackage[pagebackref=false, breaklinks=true, colorlinks, 
            citecolor=citecolor, linkcolor=linkcolor, bookmarks=false]{hyperref}
\definecolor{citecolor}{HTML}{0071BC}
\definecolor{linkcolor}{HTML}{ED1C24}

\newcommand{\figref}[1]{Figure~\ref{#1}}
\newcommand{\tabref}[1]{Table~\ref{#1}}
\newcommand{\equref}[1]{Eq.~(\ref{#1})}

\newcommand{\secref}[1]{Section~\ref{#1}}
\def\loss{\mathcal{L}\xspace}
\newcommand{\drule}{\specialrule{0.2pt}{1pt}{1pt}%
            \specialrule{0.2pt}{0pt}{\belowrulesep}%
            }

\newcommand{\ie}{\textit{i}.\textit{e}.}
\newcommand{\eg}{\textit{e}.\textit{g}.}

\definecolor{darkred}{rgb}{0.7,0.01,0.01}
\newlength\savewidth\newcommand\shline{\noalign{\global\savewidth\arrayrulewidth
  \global\arrayrulewidth 1pt}\hline\noalign{\global\arrayrulewidth\savewidth}}

\renewcommand{\paragraph}[1]{\vspace{1.mm}\noindent\textbf{#1}}

\newcolumntype{x}[1]{>{\centering\arraybackslash}p{#1pt}}
\newcolumntype{y}[1]{>{\raggedright\arraybackslash}p{#1pt}}
\newcolumntype{z}[1]{>{\raggedleft\arraybackslash}p{#1pt}}

\newcommand{\app}{\raise.17ex\hbox{$\scriptstyle\sim$}}

\definecolor{baselinecolor}{gray}{.90}
\definecolor{deemph}{gray}{0.6}

\newcommand{\gr}{\rowcolor[gray]{.90}}
\newcommand{\baseline}[1]{\cellcolor{baselinecolor}{#1}}

\title{\LARGE \bf
Test-time Adaptation in the Dynamic World \\with Compound Domain Knowledge Management
}

\author{Junha Song$^{1}$, Kwanyong Park$^{1}$, InKyu Shin$^{1}$, Sanghyun Woo$^{1}$, Chaoning Zhang$^{2}$, and In So Kweon$^{1}$
\thanks{$^{1}$J. Song, K. Pack, I. Shin, S. Woo, and I. S. Kweon are with the School of Electrical Engineering, KAIST, Daejeon 34141, Republic of Korea. {\tt\footnotesize \{sb020518, pkyong7, dlsrbgg33, shwoo93, and iskweon77\}@kaist.ac.kr}}
\thanks{$^{2}$C. Zhang is with the School of Artificial Intelligence, Kyung Hee, Yongin-si, 17104, Republic of Korea. {\tt\footnotesize chaoningzhnag1990@gmail.com}}%
}

\begin{document}

\maketitle
\thispagestyle{empty}
\pagestyle{empty}

\begin{abstract}

Prior to the deployment of robotic systems, pre-training the deep-recognition models on all potential visual cases is infeasible in practice.
Hence, test-time adaptation~(TTA) allows the model to adapt itself to novel environments and improve its performance during test time (i.e., lifelong adaptation). 
Several works for TTA have shown promising adaptation performances in continuously changing environments.
However, our investigation reveals that existing methods are vulnerable to dynamic distributional changes and often lead to overfitting of TTA models.
To address this problem, this paper first presents a robust TTA framework with compound domain knowledge management.
Our framework helps the TTA model to harvest the knowledge of multiple representative domains (\ie, compound domain) and conduct the TTA based on the compound domain knowledge. 
In addition, to prevent overfitting of the TTA model, we devise novel regularization which modulates the adaptation rates using domain-similarity between the source and the current target domain.
With the synergy of the proposed framework and regularization, we achieve consistent performance improvements in diverse TTA scenarios, especially on dynamic domain shifts.
We demonstrate the generality of proposals via extensive experiments including image classification on ImageNet-C and semantic segmentation on GTA5, C-driving, and corrupted Cityscapes datasets.
\end{abstract}

\section{INTRODUCTION}

One of the main challenges in deep neural networks is the decrease in performance during test time caused by domain shift between the pre-training domain (source) and the test domain (target)~\cite{Tsai_adaptseg_2018}.
This issue should be resolved especially for safety-critical applications like robots and autonomous vehicles, as it could pose a significant risk to application users. 
For example, in autonomous driving scenarios, failures of the recognition model on novel driving conditions may cause accidents and harm occupants.

Several research fields attempt to address this problem, such as domain generalization (DG)~\cite{gulrajani2020search, lee2021learning} and unsupervised domain adaptation (UDA)~\cite{ganin2015unsupervised}. 
In particular, UDA performs joint training using both the source and target domain during the \textit{pre-training} phase, applying techniques such as adversarial~\cite{Tsai_adaptseg_2018} or pseudo-labeling~\cite{twophase} learning. 
However, it may be challenging to utilize the UDA techniques in the following realistic cases: 
1) where the target data is unavailable, or 2) when the model is deployed on unseen domains which are different from the target domain used in pre-training.
Motivated by this issue, test-time adaptation (TTA) has been presented~\cite{tent}, assuming that the networks are pre-trained with only the source domain and the adaptation to the target domain is conducted during the \textit{testing} phase. 

\begin{figure}[t]\centering
\includegraphics[width=0.48\textwidth]{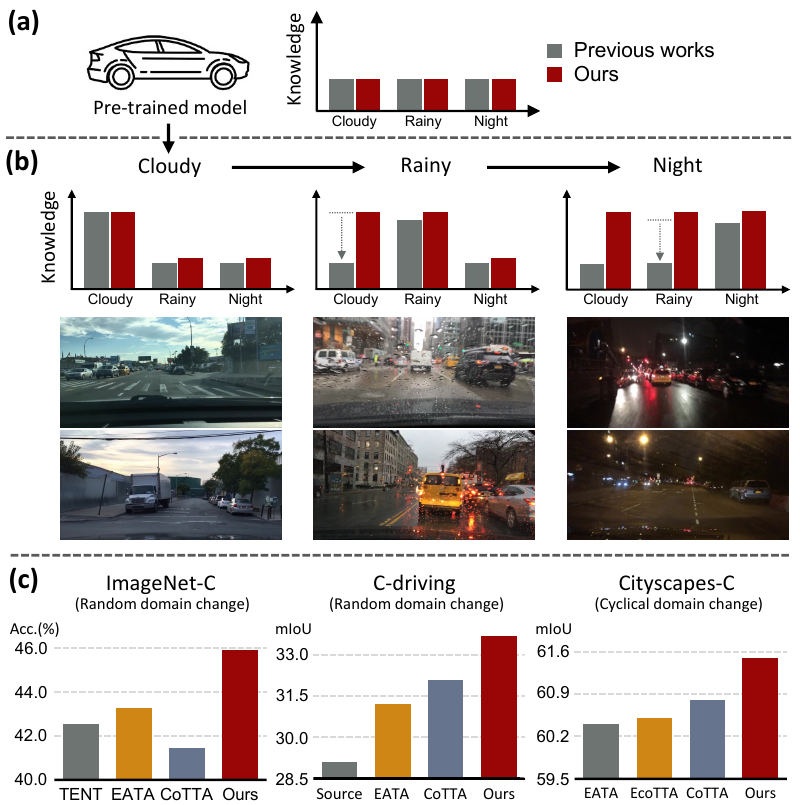}
\vspace{-2.em}
\caption{We present a conceptual illustration of our paper (a,b) and the experiment results (c). \textbf{(a) Before model deployment,} 
limited data may result in insufficient knowledge of some domains.
\textbf{(b) During Test-time adaptation (TTA),} the model can adapt to unfamiliar domains itself. Previous works only focus on adapting to the current domain while ignoring previous adaptations. On the other hand, our approach learns multiple knowledge from the compound domain, 
resulting in robust performance in diverse TTA scenarios, as shown in \textbf{the experimental results (c)}.
}
\vspace{-2.em}
\label{fig:figure1}
\end{figure}


TTA is a promising research direction for lifelong adaptation in robotic systems, which enables robots to quickly adapt to a new environment and leads to a more robust performance over time.
Due to its usefulness, TTA has garnered a great deal of attention and enjoyed a rapid performance boost by improved unsupervised loss~\cite{swr, chen2022contrastive, shin2022mm} or small-batch adaptability~\cite{wildTTA, lim2023ttn}. 
In particular, CoTTA~\cite{cotta} and EcoTTA~\cite{ecotta} presented a promising potential for lifelong adaptation by facilitating the prevention of catastrophic forgetting~\cite{eval_cotta} and error accumulation~\cite{ecotta}.
Despite these TTA breakthroughs, we have identified a notable issue with previous works. As depicted in \figref{fig:figure1}\textcolor{linkcolor}{(c)}, they demonstrate decreased performance in situations where dynamic domain changes occur, such as when encountering tunnels, sun glare, or rain while driving.
We speculate that it is because previous works barely consider managing the knowledge of multiple sub-target domains (\eg, weather or location changes), only focusing on the adaptation to the current domain while discarding the knowledge acquired from previous domains, as illustrated in \figref{fig:figure1}\textcolor{linkcolor}{(b)}.

Therefore, this paper proposes a simple yet effective approach to address this issue.
We first introduce a robust TTA framework that empowers the TTA model to learn and manage mutiple knowledge from the compound domain.
This framework utilizes our proposed continual domain-matching algorithm that efficiently discriminates the current image's domain type (\ie, pseudo-domain label).
Specifically, we continually match incoming target images with our domain prototypes and then estimate the domain type among the multiple sub-target domains. 
Moreover, we introduce a novel regularization term to prevent overfitting of the TTA model in the compound domain.
We observe that the quality of unsupervised signals is significantly varying depending on which domain the image belongs.
For example, the unsupervised signals generated by night images are often noisy, compared to the counterpart of the daytime images.
Thus, we modulate the impact of the unsupervised signals on the model by leveraging the domain-similarity between the source and the sub-target domain.

Our paper presents the following contributions:
\begin{itemize}
    \setlength\itemsep{.1em}
    \item We propose a new TTA framework that helps the TTA model to manage compound domain knowledge, which effectively copes with dynamic domain shifts.
    \item We present a domain similarity-based regularization to adjust different adaptation rates for various domain types, improving TTA performance in the wild.
    \item We evaluate our approach in diverse scenarios that can occur in the dynamic world. Our experiments consist of 1) image classification task with ImageNet-C dataset and 2) semantic segmentation task with GTA5, C-driving, and Cityscapes datasets.
\end{itemize}

\section{Related work}
\subsection{Domain Adaptation}

To mitigate the performance drop caused by the difference between the train and test domains, numerous methods for unsupervised domain adaptation (UDA)~\cite{ganin2015unsupervised, Tsai_adaptseg_2018} have been proposed. Most UDA methods alleviate the domain shift by utilizing adversarial training~\cite{advent, twophase} and self-training~\cite{cbst}. 
While the UDA methods have demonstrated their efficacy, their experimental settings had a practical limitation; the target domain was assumed to come from a single distribution (\eg, only sunny) even though the target domain can be diverged into sub-target domains (\eg, cloudy, rainy, and snowy) due to time and weather variations.
For this reason, the setup called compound domain adaptation (CDA) emerged~\cite{liu2020open, park2020discover}, taking multiple sub-target domains (\ie, compound domain) into account. 
For example, DHA-CDA~\cite{park2020discover} learns invariant representations under the compound domain by adversarially pre-training the networks with both source and multiple target datasets. 
Despite the development of DA techniques, the DA field has overlooked an important fact that the pre-trained networks can undergo the target domain which was not considered during the pre-training phase.
For instance, self-driving cars typically operate in pleasant daytime conditions, but they may also encounter unfamiliar target domains such as snow, fog, sun glare, and tunnel.

\subsection{Test-time Adaptation}

Pre-training the model with all potential target domains is difficult and expensive. 
This issue has led to the development of a new setting called as test-time adaptation (TTA), in which the adaptation is performed using the unlabeled target sample encountered by the model after deployment, regardless of the source domain on which the model was pre-trained.
TENT~\cite{tent}, a seminal work in TTA, proposes updating only small parameters (\ie, batch normalization layers) by minimizing the entropy of model predictions during test time. 
Following TENT, several TTA works improve TTA performance by designing sophisticated unsupervised loss~\cite{ttt++,swr,zhang2021memo} or improving small-batch scalability~\cite{lim2023ttn, abn}. 
Moreover, since adaptation with unsupervised loss can cause error accumulation~\cite{ecotta} and catastrophic forgetting~\cite{eval_cotta}, CoTTA~\cite{cotta} and EcoTTA~\cite{ecotta} perform continual long-term TTA under a continually changing target domain with weight restoration or regularization, respectively. 
Nevertheless, previous studies only consider domains that change in a continuous or monotonic fashion, and as a result, they tend to underperform when faced with more complex types of domain shifts, such as random or cyclical changes.
Given this issue, this paper proposes a novel solution that involves managing compound domain knowledge and regulating the model's adaptation rate for each domain separately.


\hypersetup{linkcolor=black}
\begin{figure*}[t]\centering
\includegraphics[width=0.98\textwidth]{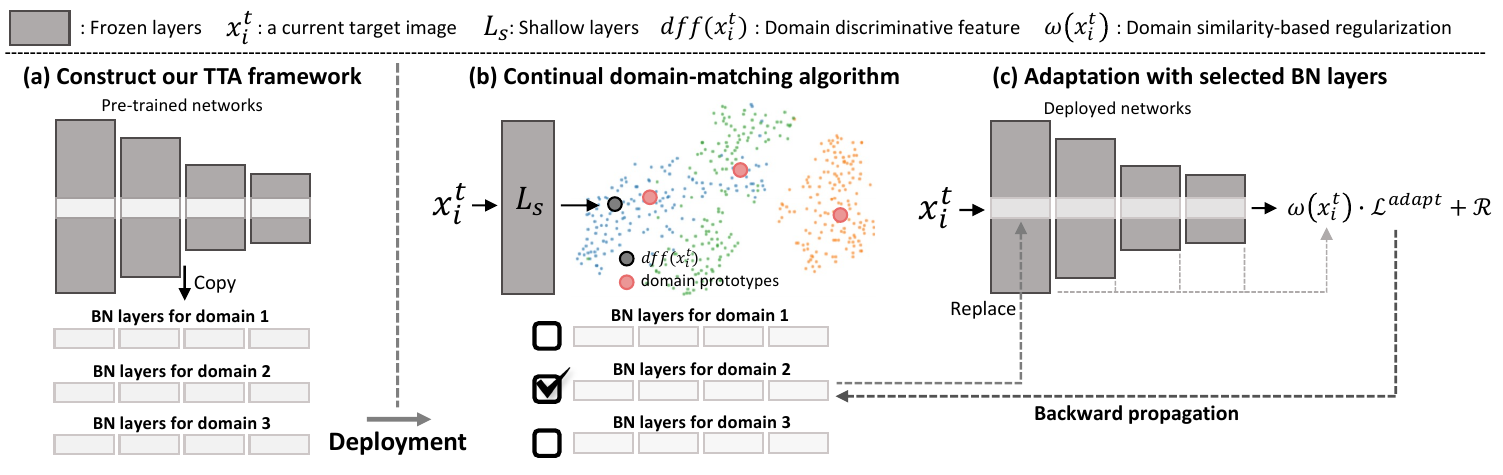}
\vspace{-1.4em}
\caption{\textbf{Overview of our approach.} (a) Before deployment, we construct a TTA framework for compound domain knowledge, cloning the batch normalization modules in the pre-trained networks. (b) In order to select one of the modules, we determine the domain type of the current target image by our proposed continual domain-matching algorithm, helping the TTA model to manage the adaptation knowledge of the compound domain. (c) We perform test-time adaptation with the selected modules, where our domain similarity-based regularization $\omega(x_i^t)$ is also applied to adjust different adaptation rates for the multiple sub-target domains. $\loss^{adapt}$ and $\mathcal{R}$ are described in \secref{sec:prerequisite}.
}
\vspace{-1.5em}
\label{fig:overview}
\end{figure*}
\hypersetup{linkcolor=linkcolor}

\section{Approach}

\subsection{Prerequisite}
\label{sec:prerequisite}

TTA models conduct the inference and adaptation simultaneously during test time to acquire more accurate prediction results in the target domain. 
However, performing TTA for a long period of time inevitably causes target domain overfitting~\cite{cotta,ecotta}, which means the model collapses and become to infers inaccurate predictions. 
This issue occurs because TTA models utilize an unsupervised loss with unlabeled target images, \ie, an unreliable backpropagation signal compared with supervised loss. 
Obviously, this issue can be avoided by adaptation early-stopping~\cite{jabbar2015methods}, but this approach is not suitable for TTA because TTA is conducted under the scenario that it is unknown when the unfamiliar target domain will emerge. 

Our work, therefore, considers long-term TTA as a basic TTA setting and leverages the adaptation loss and weight regularization proposed in EATA~\cite{eata} as a baseline to prevent overfitting. In detail about EATA, they only update the parameters $\Theta$ of batch normalization (BN) layers and freeze the rest in the networks during the adaptation process. To generate the backward propagation from $i$-th data $x_i$ in a mini-batch of $n$ target images, entropy minimization with sample filtering {\small $\loss^{adapt}$} is computed, which is formulated as:
\begin{equation}
\label{eq:ent_t}
    \loss^{adapt} =  \frac{1}{n}\sum_{i=1}^n \mathbb{I}_{\{H(\hat{y}_i) < H_0\}} 	\cdot H(\hat{y}_i),
\end{equation}

\noindent
where {\small $H(y)=-\sum_C p(y) \log p(y)$} is a formula of entropy minimization and $\mathbb{I}_{\{\cdot\}}$ is an indicator function. $\hat{y}_i$ means the logits output of $x_i$, and $H_0$, $p(\cdot)$, and $C$ refer to a pre-defined threshold, the softmax function, and the number of task classes respectively. 
Moreover, they apply the following weight regularization $\mathcal{R}$ to prevent model parameters from changing too much and thus avoid overfitting,
\vspace*{-.3em}
\begin{equation}\label{eq:weight_regularization}
    \mathcal{R}(\tilde{\Theta},\tilde{\Theta}^o)=\sum_{\theta_i\in\tilde{\Theta}} w(\theta_i) (\theta_i - \theta_i^{o})^2.
    \vspace*{-.8em}
\end{equation}
\noindent
$\tilde{\Theta}$ are adapted parameters of the networks, $\tilde{\Theta}^o$ are the corresponding frozen parameters of the pre-trained networks which are extra-stored, and $w(\theta_i)$ denotes the importance of $\theta_i$ wich is calculated by the diagonal fisher information matrix~\cite{kirkpatrick2017overcoming}. Note that our approach introduced in the next subsections desires to manage comprehensive knowledge from the compound domain or regulate the importance of adaptation loss {\small $\loss^{adapt}$}, which is orthogonal to the approach proposed in EATA and can also be applied to existing TTA works, such as TENT~\cite{tent} and CoTTA~\cite{cotta}. 

\vspace{+.3em}
\subsection{TTA Framework for compound domain knowledge}
\label{sec:cdma}

In the dynamic world, TTA models can face various environmental changes, including sudden domain shifts such as driving through tunnels and experiencing sun glare. Therefore, it is crucial for TTA models to accommodate compound domain knowledge and conduct the TTA without latency. In this subsection, we {\small (i)} present our TTA framework for managing compound domain knowledge, {\small (ii)} domain distinctive features, and {\small 
 (iii)} continual domain-matching algorithm to estimate the domain type of a current target image. \figref{fig:overview} provides the overview of our approach.



\paragraph{Compound domain TTA framework.} 
Previous works for adapting to multiple domains, such as DSBN~\cite{chang2019domain} and DSON~\cite{seo2020learning}, indicate that composing domain-specific modules helps the model to effectively learn multiple domain knowledge. 
Inspired by such a finding, we also construct $K$-specific modules consisting of BN layers to manage domain knowledge acquired from multiple sub-target domains. To be more specific, as shown in \figref{fig:overview}\,(a), we initialize the modules as cloning BN layers of the pre-trained model before deployment. During TTA, each module becomes specialized for each sub-target domain, as updating each module matched with the estimated domain type of an incoming image, as depicted in \figref{fig:overview}\,(b,c).

\paragraph{Domain distinctive features.} To interpret the domain information implied in an input image, we need to define domain-distinctive features $ddf$, a prerequisite for our domain-matching algorithm (detailed in the next paragraph). We utilize style features proposed in the field of style transfer~\cite{gatys2016image}. Specifically, convolutional feature statistics from small parts of the shallow layers $L_s$ (\eg, conv1 and layer1 in the case of ResNet) are used as the $ddf$, which can indicate the type of domain (\ie, style) of the current image~\cite{matsuura2020domain}. Consequently, our $ddf$ vectors are defined as:
\vspace*{-.2em}
\begin{equation}\label{eq:ddf}
    \resizebox{.9\hsize}{!}{
    $ddf(x_i) = \{\mu(\phi_0(x_i), \sigma(\phi_0(x_i),\mu(\phi_1(x_i), \sigma(\phi_1(x_i))\}$,
    }
    \vspace*{-.2em}
\end{equation}

\noindent
where $\mu$ and $\sigma$ denotes the feature statistics and $\phi_0$ and $\phi_1$ are the final output of the conv1 and layer1, respectively.

\begin{figure}[t]\centering
\includegraphics[width=0.49\textwidth]{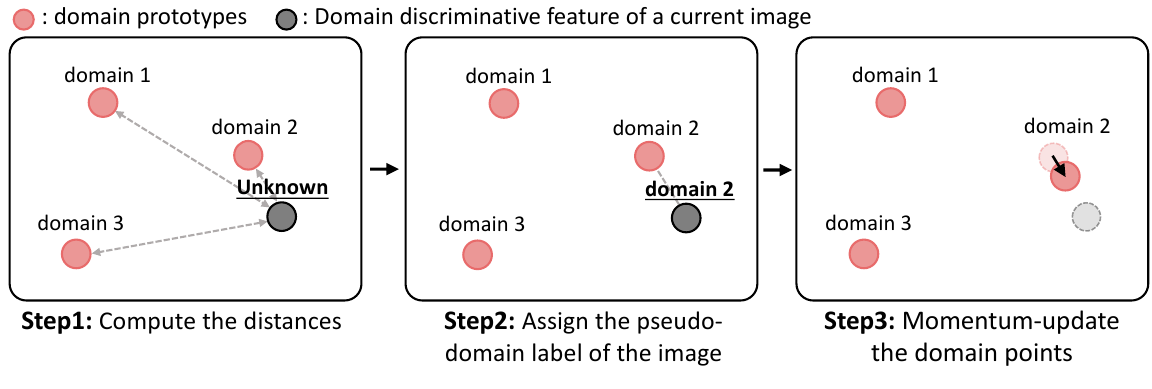}
\vspace{-2.6em}
\caption{\textbf{Continual domain-matching algorithm} Our fast matching algorithm allows the model to effectively predict the pseudo domain label of target images. Through this process, each domain prototype becomes representative of each type of sub-domains.}
\vspace{-1.8em}
\label{fig:cdma}
\end{figure}

\paragraph{Continual domain-matching algorithm.} With the $ddf$ defined, we need a fast way to determine the pseudo-domain label to make $K$-specific modules optimized to each sub-target domain.
As proposed in \cite{matsuura2020domain}, kmeans clustering~\cite{kmeans} can be one of the potential solutions, but we note that such iterative clustering algorithms require long running time when we deal with enormous vectors (\ie, all $ddf$ vectors of incoming target samples). Instead of the clustering algorithm, we focus on an effective matching algorithm and thus propose a continual domain-matching algorithm. Our algorithm,\,visualized in\,\figref{fig:cdma} and defined as the following, is continuously executed as the target image incomes:

\begin{enumerate}
    \item Compute the distances between the domain prototypes {\small $\{d_{k}\}_{k=1}^{K}$} and the $ddf$ of the current image $x_i$. 
    \item Assign the index {\small $l$} of the nearest domain prototype {\small $d_{l}$} as the pseudo domain label for the $x_i$. 
    \item Update the selected domain prototype by moving average technique formulated as {\small $d_{l} \gets \eta \cdot d_{l} + (1-\eta) \cdot ddf(x_i)$}, where {\small $\eta$} represents momentum.
    \item Conduct test-time adaptation with the $l$-th module, as shown in \figref{fig:overview}~(c).
\end{enumerate}

\noindent
Specifically, the distance function is used as the Bhattacharya function~\cite{kailath1967divergence} which is useful for comparing two statistics $(p, q)$ (as analyzed in \secref{sec:ablation}) and formulated as:
\vspace{-.2em}
\begin{equation}\label{eq:bhata}
  \resizebox{.9\hsize}{!}{
  $D_{B}(p, q)=\frac{1}{4} \ln \left(\frac{1}{4}\left(\frac{\sigma_{p}^{2}}{\sigma_{q}^{2}}+\frac{\sigma_{q}^{2}}{\sigma_{p}^{2}}+2\right)\right)+\frac{1}{4}\left(\frac{\left(\mu_{p}-\mu_{q}\right)^{2}}{\sigma_{p}^{2}+\sigma_{q}^{2}}\right)$.
  }
\end{equation}

\noindent
Through our proposed algorithm, each domain prototype is derived to represent each sub-target domain. In addition, the domain prototypes are initialized before model deployment by utilizing kmeans clustering with the $ddf$s of the source dataset. We highlight that access to the source data prior to deployment is possible in cases where the source dataset is publicly accessible or if the owner of a pre-trained model attempts to adapt the model to a target domain, as stated in previous TTA works~\cite{ttt++, ecotta, eata}. 

\subsection{Domain Similarity-based Regularization.}
TTA models adapt to new domains using \textit{unsupervised} loss (\eg, entropy minimization~\cite{tent}) since \textit{unlabeled} target images are typically collected in edge devices. Unfortunately, the unsupervised loss is less reliable than the supervised loss, occasionally containing false backpropagation signals when the calculated loss comes from wrong predictions. Existing TTA works, such as EATA~\cite{eata} and EcoTTA~\cite{ecotta}, demonstrate that this issue leads to error accumulation and model failure (\ie, overfitting).

We observe that the pace of error accumulation varies according to the sub-target domain since images gathered in the compound domain make different qualities of unsupervised loss. For example, night images typically provide less accurate predictions than daytime images and thus the night-specific module becomes quickly collapsed (\ie, overfitted). 
Therefore, we attempt to alleviate error accumulation by regularizing the adaptation pace, especially when an unsupervised loss is likely to be more unreliable since the sub-target domain is significantly distinct from the pre-training domain.
Specifically, we compute a regularization term $\omega$ leveraging the similarity between a target image's statistics and source statistics $\mathcal{S}$ in BN layers (\ie, running estimates~\cite{adapt_abn}), which is defined as:
\vspace{-.3em}
\begin{equation}
\label{eq:reg_weight}
    \resizebox{.8\hsize}{!}{
    $\omega(x_i) =\mathbb{E}_{l \sim L} \: \text{sim} \{ \, (\mu(\phi_l(x_i), \sigma(\phi_l(x_i))) \:,\: \mathcal{S}_l \, \}^{\gamma}$,
    }
\end{equation}

\noindent
where we use cosine similarity, $L$ denotes last BN layers in layer1${\sim}$layer4 in the case of ResNet backbone~\cite{pytorch}. The scale parameter $\gamma$ controls the strength of the regularization. Overall, the total loss $\loss^{total}$ for test-time adaptation to the target domain is formulated as: 
\begin{equation}
\label{eq:totalloss}
    \loss^{total} = \omega(x_i) \cdot \loss^{adapt} + R.
\end{equation}

\section{Experimental Results}

In this section, we present and verify the effectiveness of our approach on various benchmarks, including image classification and semantic segmentation.

\vspace{+.3em}
\subsection{Classification Experiments}

{\renewcommand{\arraystretch}{1.25}
\begin{table}[b]
\vspace{-1.em}
\caption{\textbf{Comparision on ImageNet-C.} We report the average error rate (\%) over 10 \underline{diverse corruption sequences}. The wall-clock time refers to the total time taken to adapt to one sequence composed of 15 corruptions in ImageNet-C. Source denotes the pre-trained model without adaptation.}
\vspace{-3.2em}
\begin{center}
\large{\resizebox{0.49\textwidth}{!}{ 
\begin{tabular}{l cccccc}
\hline
ImageNet-C      & Source   & TENT~\cite{tent} & TTT+~\cite{ttt++}     & EATA~\cite{eata}       & CoTTA~\cite{cotta}               & \baseline{Ours}        \\
\drule
Avg.\,error\,{\small $\downarrow$} & 74.5\,(0) & 56.1\,(0.6)&  58.2\,(0.8) & 54.9\,(2.3) & 54.6\,(3.9) & \baseline{\textbf{53.5\,(1.6)}}  \\
Wall\,time\,{\small $\downarrow$}      & 4m 35s   & 7m 18s    &  8m 12s      & 7m 23s     & \textcolor{darkred}{22m 37s}    & \baseline{7m 52s}     \\
\hline
\end{tabular}}}
\end{center}
\vspace{-.8em}
\label{tab:imagenet_avg}
\end{table}}
{\renewcommand{\arraystretch}{1.2}
\begin{table*}[t]
\caption{\textbf{Comparision on two notable sequences.} We conduct experiments on continual TTA setup. Avg. err means the average error rate (\%) of all 15 corruptions. Two below Sequences (in short, seq.) represent situations in which 15 corruptions of the four main categories, \ie, \textbf{N}oise, \textbf{B}lur, \textbf{W}eather, and \textbf{D}igital, occur in a \underline{continuous} and \underline{random} manner, respectively.}
\vspace{-2.0em}
\begin{center}
\large{\resizebox{0.99\textwidth}{!}{ 
\begin{tabular}{lcccccccccccccccc}
Time       & \multicolumn{15}{c}{$t\xrightarrow{\hspace*{25cm}}$}  &    \\[-.1em]
\hline
\textbf{Continuous seq.} & Gaus.\,N & Shot.\,N & Impu.\,N & Defo.\,B & Glas.\,B & Moti.\,B & Zoom\,B & Snow\,W  & Fros.\,W & Fog.\,W  & Brig.\,W & Cont.\,D & Elas.\,D & Pixe.\,D & Jpeg\,D  & Avg.\,err\,{\small $\downarrow$}  \\
\drule
Source     & 94.3    & 88.5    & 94.6    & 74.3    & 82.9    & 71.0    & 64.7   & 76.8    & 72.2    & 76.1    & 49.2    & 84.3    & 71.4    & 65.2    & 52.2    & 74.5        \\
TENT~\cite{tent} & 61.6 & 58.9 & 56.8 & 59.5 & 61.6 & 56.0 & 53.9 & 56.9 & 59.6 & 52.4 & 45.2 & 66.3 & 52.9 & 53.9 & 55.9 & 56.8  \\
TTT++~\cite{ttt++}     & 63.8 & 60.4 & 58.1 & 60.7 & 64.1 & 55.6 & 51.8 & 57.7 & 59.2 & 51.0 & 49.5 & 65.9 & 55.1 & 58.7 & 57.2 & 57.9  \\
EATA~\cite{eata}       & \textbf{61.1}    & 57.1    & 53.2    & \textbf{55.6}    & 59.6    & 54.2    & \textbf{51.6}   & 55.9    & 56.7    & 51.1    & 45.4    & \textbf{52.9}    & 49.1    & 49.2    & 49.8    & 53.5        \\
CoTTA~\cite{cotta}      & 61.9    & 58.3    & 54.6    & 57.3    & 59.8    & 55.2    & 53.6   & 56.2    & \textbf{56.6}    & 51.3    & 49.6    & 53.3    & 50.6    & 51.3    & \textbf{49.3}    & 54.6        \\
\gr
Ours       & 61.2    & \textbf{56.4}    & \textbf{52.6}    & 56.0    & \textbf{59.4}    & \textbf{53.3}    & 52.4   & \textbf{54.9}    & 56.8    & \textbf{50.5}    & \textbf{45.1}    & 53.2    & \textbf{48.6}    & \textbf{49.0}    & 50.2    & \textbf{53.3}        \\
\hline
\\[-.3em]
\hline
\textbf{Random seq.} & Cont.\,D & Gaus.\,N & Defo.\,B & Zoom\,B  & Fog.\,W  & Glas.\,B & Jpeg\,D & Fros.\,W & Pixe.\,D & Elas.\,D & Shot.\,N & Impu.\,N & Snow\,W  & Moti.\,B & Brig.\,W & Avg.\,err\,{\small $\downarrow$}  \\
\drule
Source     & 84.3    & 94.3    & 74.3    & 64.7    & 76.1    & 82.9    & 52.2   & 72.2    & 65.2    & 71.4    & 88.5    & 94.6    & 76.8    & 71.0    & 49.2    & 74.5        \\
TENT~\cite{tent} & 56.4 & 60.6 & 62.6 & \textbf{52.2} & 57.2 & 62.4 & 50.8 & 60.8 & 54.8 & 51.0 & 62.5 & 59.1 & 61.9 & 59.8 & 54.0 & 57.5  \\
TTT++~\cite{ttt++}     & 60.2 & 62.9 & 63.6 & 52.0 & 58.5 & 64.8 & 55.8 & 62.6 & 56.2 & 51.3 & 64.7 & 61.7 & 60.4 & 61.2 & 58.3 & 59.6  \\
EATA~\cite{eata}       & 54.9    & 66.4    & 62.2    & 57.0    & 58.9    & 62.4    & 52.4   & 57.9    & 52.1    & 50.3    & 57.6    & 57.2    & 57.7    & 55.8    & 48.5    & \textcolor{darkred}{56.8}        \\
CoTTA~\cite{cotta}      & 59.4    & 68.1    & 64.6    & 63.3    & 61.6    & 62.8    & 53.7   & 57.8    & 51.6    & 53.4    & 57.2    & 57.0    & 58.6    & 57.5    & 51.3    & \textcolor{darkred}{58.5}        \\
\gr
Ours       & \textbf{53.9}    & \textbf{62.0}    & \textbf{55.8}    & 53.1    & \textbf{52.2}    & \textbf{61.0}    & \textbf{51.2}   & \textbf{56.8}    & \textbf{49.7}    & \textbf{48.1}    & \textbf{56.1}    & \textbf{54.0}    & \textbf{55.7}    & \textbf{54.0}    & \textbf{47.6}    & \textbf{54.1}       \\
\hline
\end{tabular}}

}\end{center}
\vspace{-2.6em}
\label{tab:imagenet_case}
\end{table*}}

\paragraph{Experimental setup.} We use ImageNet and ImageNet-C dataset~\cite{imagenetC} as the source (\ie, pre-training) and the target (\ie, adaptation) domain, respectively. The ImageNet-C includes 15 diverse sub-target types of 4 main categories (\ie, Noise, Blur, Weather, and Digital). Following the previous TTA studies~\cite{eata, cotta}, we conduct experiments in a continual TTA setup, where we continually adapt the deployed model to each corruption type sequentially without resetting the model. Specifically, by using the official code of CoTTA~\cite{cotta}, evaluations are executed on the continual TTA setup with ten diverse corruption-type sequences, which provide a more comprehensive evaluation of TTA methods. We emphasize that diverse sequences include both the continuous flow as well as random shifts in corruption types.

\paragraph{Implementation Details.} We use the ResNet-50 pre-trained with AugMix~\cite{augmix}, which is officially provided in RobustBench~\cite{robustbench}. After deployment, we utilize the SGD optimizer, the learning rate of 5e-3, the batch size of 32, and the same hyperparameters for adaptation loss $\loss^{adapt}$ and weight regularization $\mathcal{R}$ as EATA~\cite{eata} (such as $H_0$ of $0.4 {\times} \ln C$, where $C$ is the number of task classes). We set the number of specific modules $K$, the scale parameter $\gamma$, the momentum $\eta$, to 3, 1.5, and 0.9, respectively. Their ablation study can be found in \secref{sec:ablation}.

\paragraph{Comparison with TTA methods.} We report the average error rate (\%) of ten diverse sequences in \tabref{tab:imagenet_avg} and we detail two notable sequences among the ten sequences in \tabref{tab:imagenet_case}. The first sequence means a scenario where corruptions belonging to the same main category \textit{continuously} appear, whereas the second sequence depicts a situation where all corruption types are \textit{randomly} distributed. The results demonstrate that our technique achieves robust TTA performance even with random domain change, compared to competing TTA methods. To be more specific, our approach shows similar performance regardless of continuous and random changing. In contrast, CoTTA~\cite{cotta} and EATA~\cite{eata} show significantly degraded performance in the random domain change. We suggest that separating and managing the compound domain knowledge enables our approach to adapt widely to all corruption types and perform well even in randomly changing domains. 

In addition, we measure the wall-clock time for each TTA method to adapt to all 15 corruptions as presented in \tabref{tab:imagenet_avg}. Our approach requires only a negligible overhead compared to CoTTA. Furthermore, we evaluate the efficiency of our domain-matching algorithm against the kmeans clustering. To adapt to only one corruption (i.e., 5000 samples), our TTA approach takes 32s\,(156fps), whereas it takes 28m\,16s\,(3fps) when we simply replace our domain-matching algorithm with the kmeans clustering. This suggests that the iterative clustering algorithm may not be suitable for TTA in practice.

\begin{figure}[b]\centering
\vspace{-1.3em}
\includegraphics[width=0.49\textwidth]{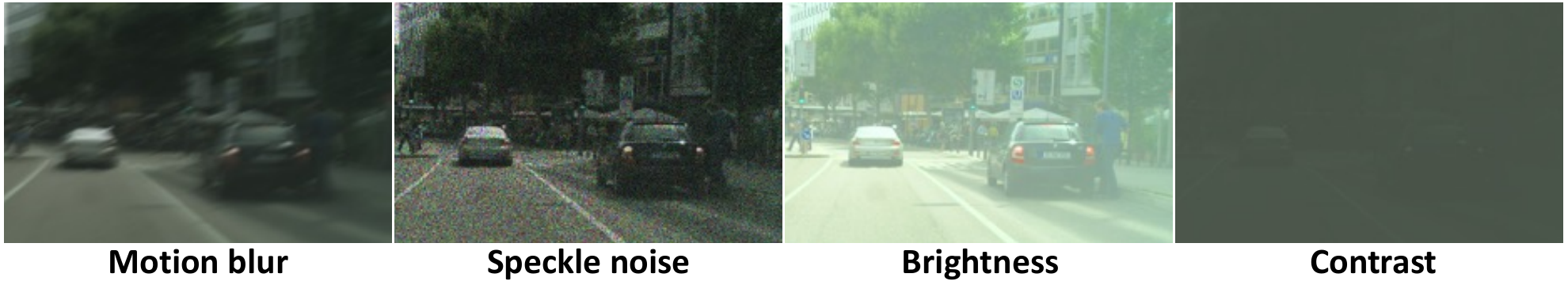}
\vspace{-2.em}
\caption{\textbf{Examples of Cityscapes with corruption.} We evaluate our approach under the scenario where the four types of corruption cyclically come across. The experiment results can be found in \tabref{tab:cityscapes_c}.}
\label{fig:cityscapesC}
\end{figure}

\vspace{+.3em}
\subsection{Segmentation Experiments}

{\renewcommand{\arraystretch}{1.2}
\begin{table}[b]
\vspace{-1.5em}
\caption{\textbf{Comparision on C-driving dataset.} 
The average mIoU score for six domain sequences, created by \underline{randomly} mixing three sub-attributes, is reported for GTA5 to C-driving experiment.}
\vspace{-3.0em}
\begin{center}
\large{\resizebox{0.49\textwidth}{!}{ 
\begin{tabular}{l|ccc x{32}|ccc x{32}}
\hline
Random seq.  & \multicolumn{4}{c|}{Time-of-day}   & \multicolumn{4}{c}{Weather}     \\
mIoU\,{\small $\uparrow$}       & daytime & night & twilight & Mean  & cloudy & rainy & snowy & Mean   \\
\drule
Source       & 29.6    & 11.2  & 26.1     & 22.3 & 32.8   & 27.0   & 27.5  & 29.1  \\
TENT~\cite{tent}    & 18.9    & 10.5  & 17.4     & 15.6 & 23.9   & 17.4   & 22.0  & 21.1  \\
EATA~\cite{eata}         & 30.8    & 15.1  & 29.4     & 25.1 & 35.5   & 29.1   & 29.0  & 31.2  \\
CoTTA~\cite{cotta}        & 32.1    & 15.4  & 30.2     & 25.9 & 36.4   & 29.6   & 30.2  & 32.1  \\
\gr
Ours w/o $\omega(x_i)$ & 33.3    & 15.6  & 32.3     & 27.0 & 37.1   & 30.8   & 31.8  & 33.2  \\
\gr
Ours         & \textbf{33.6}    & \textbf{16.5}  & \textbf{32.7}     & \textbf{27.6} & \textbf{37.5}   & \textbf{31.2}   & \textbf{32.3}  & \textbf{33.7}  \\
\hline
\end{tabular}
}}\end{center}
\vspace{-1.0em}
\label{tab:gtav}
\end{table}}

{\renewcommand{\arraystretch}{1.2}
\begin{table*}[t]
\vspace{-1.0em}
\caption{\textbf{Comparision on the corrupted Cityscapes dataset.} We simulate the \underline{cyclical} driving conditions encountered in the real world. The results show if the TTA method is able to adapt over a long period of time to multiple sub-target domains.}
\vspace{-2.0em}
\begin{center}
\large{\resizebox{0.99\textwidth}{!}{ 
\begin{tabular}{l |cccc|cccc|cccc|cccc| x{40}}
\multicolumn{1}{l}{}      & \multicolumn{17}{l}{$t\xrightarrow{\hspace*{26.6cm}}$}  \\[-.1em]
\hline
Round     & 1       &         &         &         & 4       &         &         &         & 7       &         &         &         & 10      &         &         &         &   mIoU\,{\small $\uparrow$}    \\
Domain    & Moti.\,B & Spec.\,N & Brig.\,W & Cont.\,D & Moti.\,B & Spec.\,N & Brig.\,W & Cont.\,D & Moti.\,B & Spec.\,N & Brig.\,W & Cont.\,D & Moti.\,B & Spec.\,N & Brig.\,W & Cont.\,D & Mean   \\
\drule
Source    & 31.3    & 11.5    & 60.9    & 27.3    & 31.3    & 11.5    & 60.9    & 27.3    & 31.3    & 11.5    & 60.9    & 27.3    & 31.3    & 11.5    & 60.9    & 27.3    & 32.8  \\
TENT~\cite{tent} & 57.9    & 53.2    & 69.1    & 59.8    & 54.4    & 49.5    & 63.9    & 54.1    & 44.1    & 39.9    & 52.0      & 42.9    & 37.7    & 34.2    & 44.0      & 36.3    & 49.6  \\
EATA~\cite{eata}      & \textbf{58.1}    & 53.5    & 69.6    & 60.7    & 57.5    & 53.5    & 69.6    & 60.7    & 57.5    & 53.5    & 69.6    & 60.7    & 57.5    & 53.5    & 69.6    & 60.7    & 60.4  \\
EcoTTA~\cite{ecotta}    & \textbf{58.1}    & 53.5    & 69.8    & 60.7    & 57.8    & 53.5    & 69.8    & 60.7    & 57.8    & 53.5    & 69.8    & 60.7    & 57.8    & 53.5    & 69.8    & 60.7    & 60.5  \\
CoTTA~\cite{cotta}     & 57.9    & 53.6    & 70.8    & 61.2    & 57.9    & 53.4    & 70.8    & 61.0    & 57.8    & 53.4    & 70.8    & 61.0    & 57.8    & 53.4    & 70.8    & 61.2    & 60.8  \\
\gr
Ours      & 58.0    & \textbf{54.5}    & \textbf{71.6}    & \textbf{62.0}    & \textbf{57.9}    & \textbf{54.3}    & \textbf{71.8}    & \textbf{62.2}    & \textbf{57.9}    & \textbf{54.3}    & \textbf{71.8}    & \textbf{62.2}    & \textbf{57.9}    & \textbf{54.3}    & \textbf{71.8}    & \textbf{62.2}    & \textbf{61.5}  \\
\hline
\end{tabular}
}}\end{center}
\vspace{-.6em}
\label{tab:cityscapes_c}
\end{table*}}

\begin{figure*}[t]\centering
\vspace{-0.9em}
\includegraphics[width=0.98\textwidth]{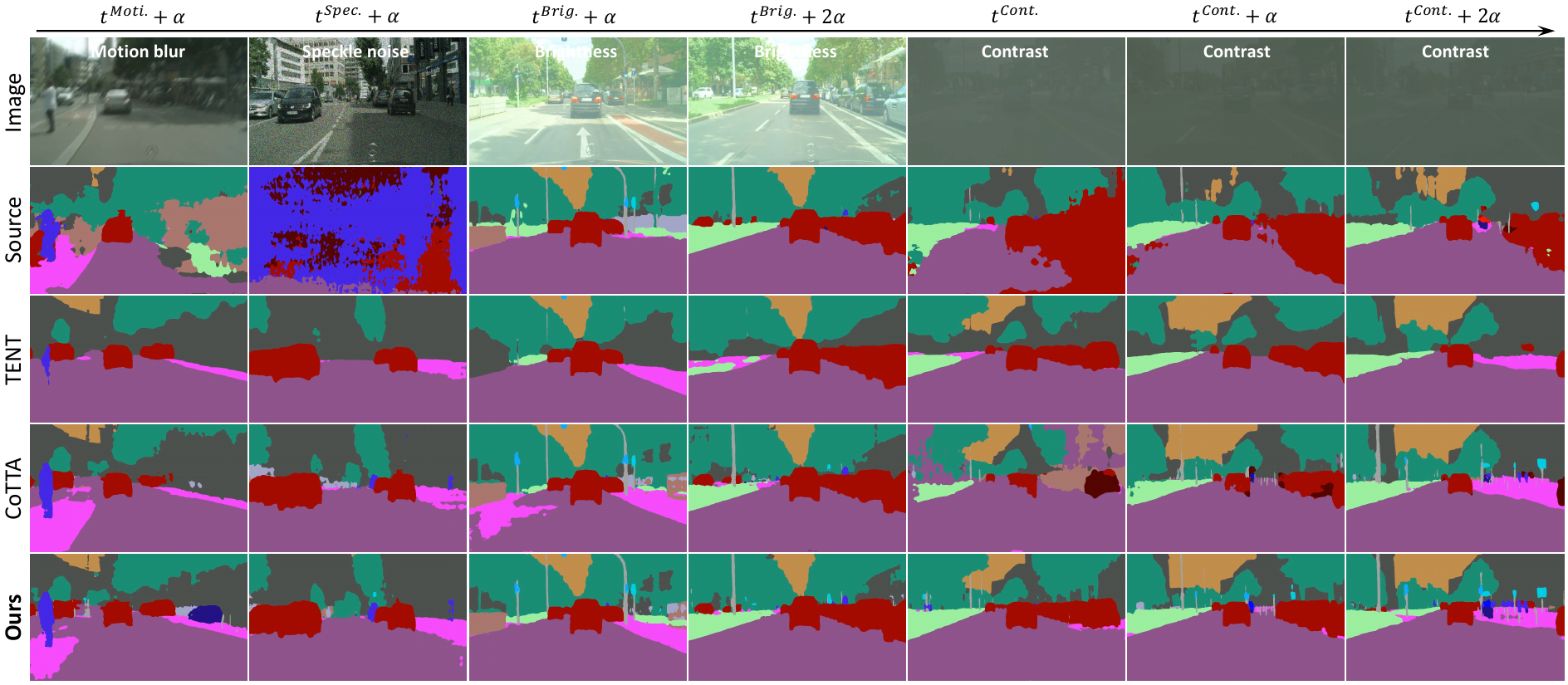}
\vspace{-0.9em}
\caption{\textbf{Qualitative comparison with TTA methods.} We visualize the prediction results in the fourth round. $t^{X}$ denotes the moment samples corrupted by $X$ type come in and $\alpha$ represents a short period of time (\eg, one second). The results of Source and TENT~\cite{tent} reflect the problem of domain shift and overfitting to the target domain, respectively. In comparison to CoTTA~\cite{cotta}, our approach shows robust performance even in dynamic domain change.}
\vspace{-1.8em}
\label{fig:qualitative}
\end{figure*}

\paragraph{Experimental setup.} 
We conduct our experiments with \{GTA5, C-driving\} and \{Cityscapes, Cityscapes with corruption\} where the former and latter dataset represents the source and the target domain, respectively. 
The GTA5~\cite{gta5} dataset contains synthetic images from the video game, Grand Theft Auto 5, whereas the C-driving dataset~\cite{liu2020open} includes the real-world samples which are separated by weather attributes (\ie, cloudy, rainy, and snowy) or time-of-day attributes (\ie, daytime, twilight, and night). 
The Cityscapes with corruption~\cite{swr} (Cityscapes-C) are created by applying the 4 types of corruption selected from each of the main corruption categories to the validation set of Cityscapes~\cite{cityscapes}; we use motion blur, speckle noise {\small (close to sensor noise)}, brightness~{\small (close to sun glare)}, and contrast~{\small(close to tunnel)}, which are illustrated in \figref{fig:cityscapesC}. 
Following the experimental setup of CoTTA~\cite{cotta}, we repeat the same sequence group (\eg, weather or time-of-day attributes of C-driving dataset and four types of corrupted Cityscapes) 10 times so that we mimic the real-life scenario where driving environments might be revisited. 
It should be noted that this setup also shows the adaptation performance in the long term. 
Additionally, when the target is the C-driving dataset, we utilize 400 unlabeled images from the train set of each sub-attribute for the adaptation, as suggested in CoTTA.

\paragraph{Implementation Details.} 
The source models for GTA5 and Cityscapes datasets are respectively based on DeepLabV3 and DeepLabV3+ with ResNet-50.
The GTA5 model is pre-trained according to the official code and specifications of CBST~\cite{cbst}, while the Cityscapes model is used as the publicly-available one from RobustNet \cite{robusetnet}. 
For adaptation to the target domain, we use the batch size of 2, the SGD optimizer, and the learning rate of 1e-5. The image size of the C-driving and Cityscapes datasets is set to 1280${\times}$720 and 1440${\times}$720, respectively. 
As mentioned in \secref{sec:prerequisite}, our method can be utilized with exisit TTA works; therefore, for semantic segmentation tasks, we evaluate our approach using the adaptation loss of CoTTA~\cite{cotta} which is cross entropy loss using the pseudo label with multi-scaling input (scales=[0.75, 1.0, 1.25, 1.5]) and flip. 
The rest of the implementation details are the same as for image classification.  

\paragraph{Comparison with TTA methods.} 
In the experiment of the GTA5 to C-driving dataset, we report the average mIoU (\ie, mean intersection over union) score using six domain sequences made by \textit{randomly} mixing the order of three sub-attributes. As shown in \tabref{tab:gtav}, our approach achieves the best performance among the TTA works. We also describe the adaptation performance of our approach without our regularization term $\omega(x_t)$. The results indicate that our regularization contributes to improved performance, especially in domains that are challenging to generate a high-quality unsupervised loss, such as night, rainy, and snowy conditions.

We report the mIoU score of four types of corrupted Cityscapes across 10 \textit{cyclical} rounds, as described in \tabref{tab:cityscapes_c}. Also, we provide qualitative comparisons as shown in \figref{fig:qualitative}, where the prediction results in the \textit{fourth} round are visualized using Cityscape video demo~\cite{cityscapes}. 
Our approach outperforms existing TTA approaches by acquiring comprehensive knowledge from the compound domain. TENT~\cite{tent} suffers from a gradual decline in performance because it does not attempt to alleviate the overfitting problem in long-term adaptation. Even though CoTTA~\cite{cotta} shows consistent performance over time, their performance is inferior to ours. \figref{fig:qualitative} shows the problem that CoTTA fails to retain the acquired knowledge from cyclical sub-target domains. Specifically, although CoTTA adapted to the contrast domain (close to tunnel and underground) in the previous rounds, once the contrast domain re-encounters (\ie, $t^{cont.}$), the CoTTA model predicts unreliable results since the model may be strongly optimized for the last domain, brightness (close to daytime and sun glare). 
On the other hand, our approach makes stable predictions even in abruptly changing domains. We believe that the adaptation knowledge across multiple sub-target domains is effectively stored and managed by our TTA framework for compound domain knowledge, allowing our method to outperform its competitors.

\vspace{+.6em}
\section{Empirical Study}
\label{sec:ablation}

\begin{figure}[t]\centering
\includegraphics[width=0.46\textwidth]{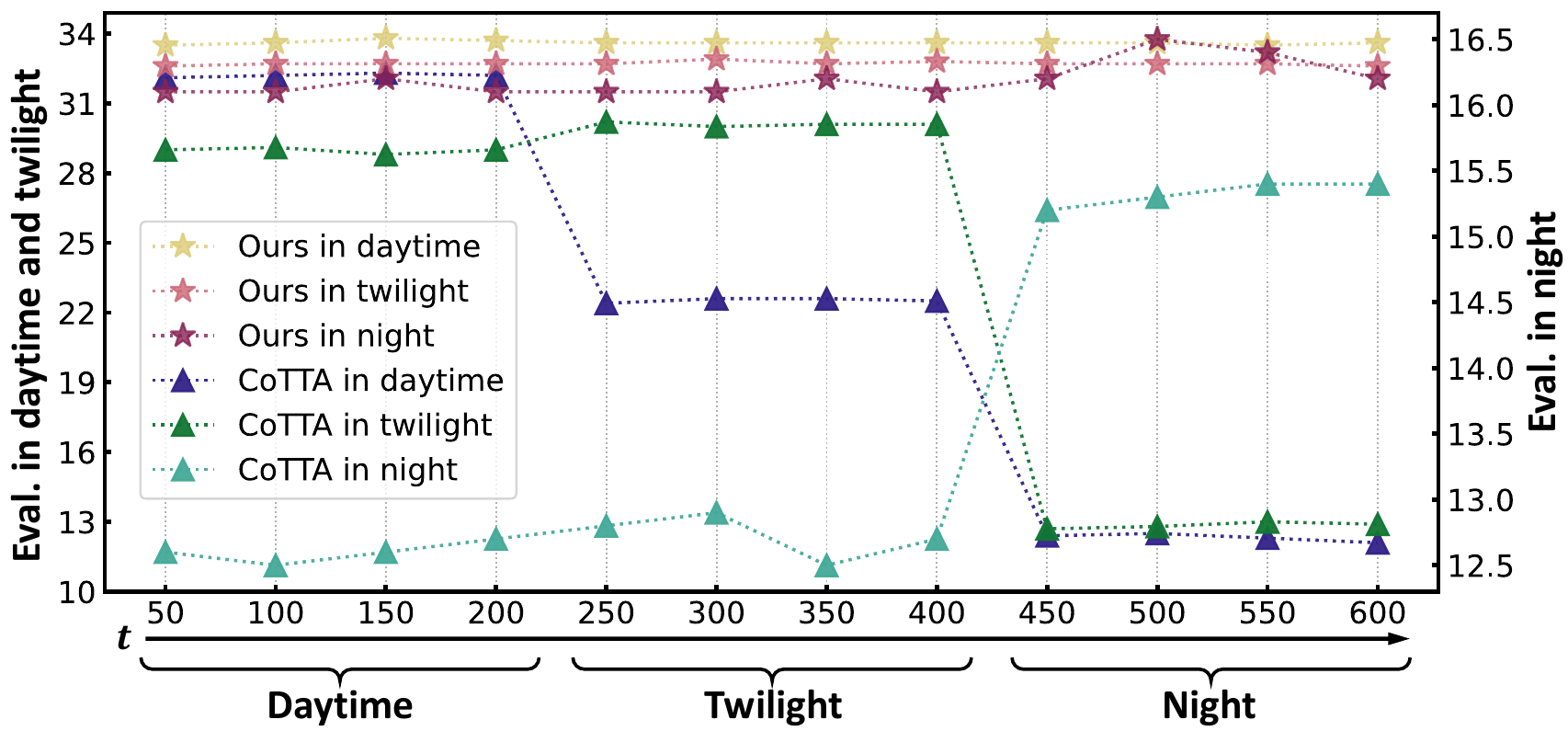}
\vspace{-1.em}
\caption{\textbf{TTA with compound domain knowledge management.} We freeze the TTA model every 50 iterations in the fourth round and calculate the mIoU score for each time-of-day attribute of the C-driving dataset. The x-axis represents the number of iterations and the incoming domain, and the y-axis refers to the evaluation results of the frozen model for each attribute.}
\vspace{-.5em}
\label{fig:compound}
\end{figure}

\paragraph{Compound domain knowledge management} Provided the unexpected domain shift and the situation that the TTA model can not perform TTA due to a sudden lack of computational resources, we need to allow the model to learn comprehensive knowledge and guarantee a consistent performance in the compound domain. As shown in \figref{fig:compound}, once finishing adaptation to every 50 iterations, we freeze the model and conduct evaluation on diverse environments, where we utilize time-of-day attributes of the C-driving dataset. Despite of the time flow and domain change, our approach shows robust performance thanks to compound domain knowledge managed in our proposed framework. In contrast, CoTTA only focuses on adapting to the current domain (\ie, narrow knowledge), resulting in significant performance degradation in different domains. This issue may pose a risk to application users in unexpected events.

\begin{figure}[t]\centering
\includegraphics[width=0.46\textwidth]{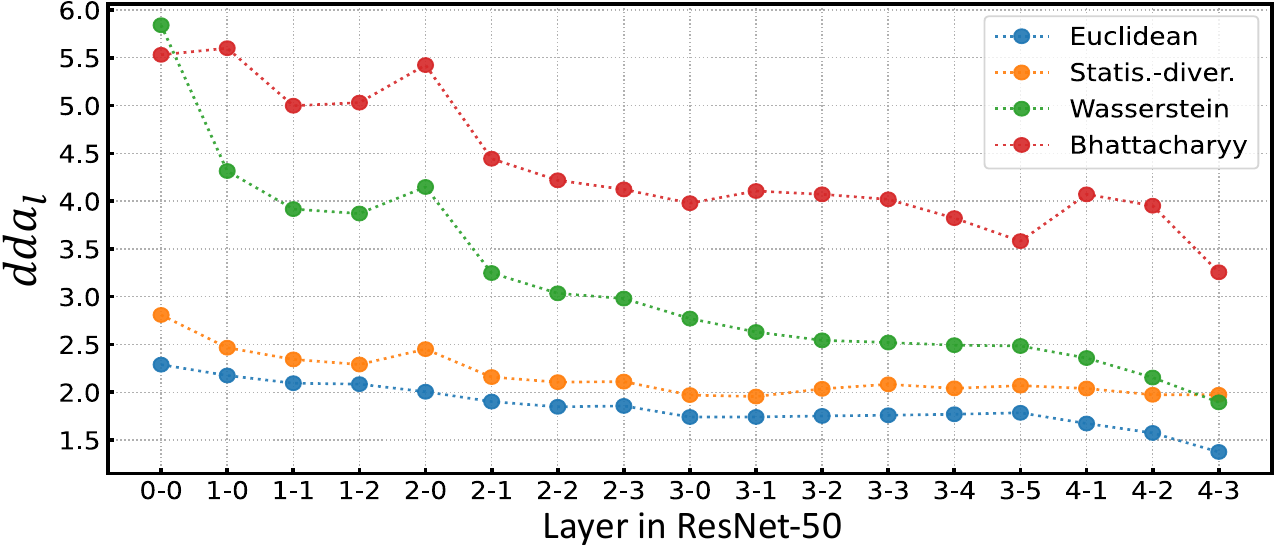}
\vspace{-1.em}
\caption{\textbf{Domain discrimination ability of distance functions.} We conduct experiment with ResNet-50~\cite{pytorch} on the C-driving dataset. On the x-axis, the $n_{1}{-}n_{2}$ means the $n_2$-th block of layer-$n_1$ in the case of ResNet backbone.
The $dda_l$ on the y-axis denotes domain discrimination ability of the function in $l$-layer, which is defined as \equref{eq:sdr}.}
\vspace{-2.em}
\label{fig:dr}
\end{figure}

\paragraph{Number of domain-specific modules~{\small $K$}.}
To analogy the impact of {\small $K$}, we conduct the following experiment on both time-of-day and weather attributes of the C-driving dataset. 1) Assume that the target images have already been collected. 2) Assign domain labels to the images using kmeans clustering \cite{matsuura2020domain}. 3) Using the estimated domain labels, train the {\small $K$} specific modules. The results are shown in the following.
{\renewcommand{\arraystretch}{1.1}
\begin{table}[h]
\vspace{-2.5em}
\begin{center}
\large{\resizebox{0.30 \textwidth}{!}{ 
\begin{tabular}{lccccc}
K    & 1    & 2    & 3    & 4    & 5     \\
\shline
mIoU\,{\small $\uparrow$} & 38.6 & 39.6 & \textbf{40.2} & 39.8 & 39.2  \\
\end{tabular}
}}\end{center}
\vspace{-2.5em}
\label{tab:oracle}
\end{table}}

\begin{figure}[t]\centering
\includegraphics[width=0.48\textwidth]{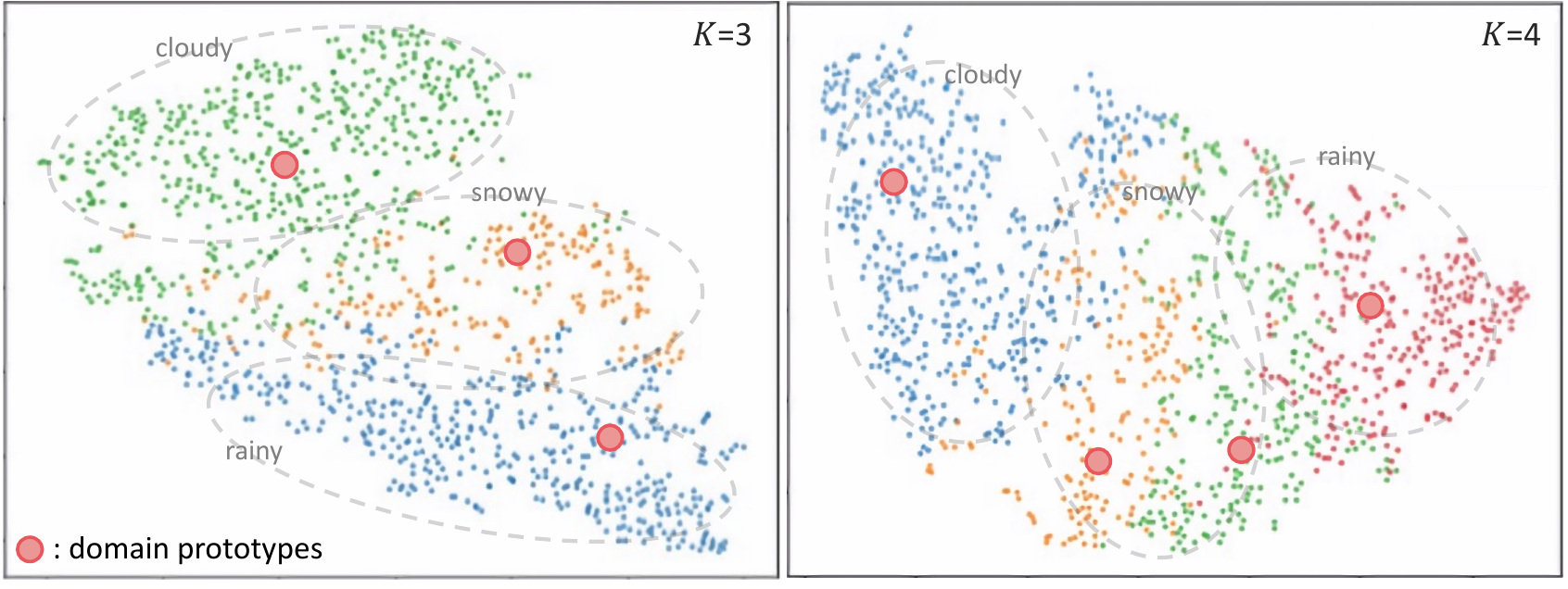}
\vspace{-1.9em}
\caption{\textbf{Example results of continual domain-matching algorithm.} On whether attributes of the C-driving dataset, we visualize the effect of our continual domain-matching algorithm. After the domain prototypes are specialized to each domain in the first round, we scatter $ddf(x^{t}_{i})$ of the images in the second round according to the matched domain which is represented as different colors. We use t-SNE~\cite{maaten2008visualizing} for visual evaluation. The gray text indicates the label for the oracle domain.}
\vspace{-1.8em}
\label{fig:clustering_vis}
\end{figure}

\noindent
We observe that constructing experts for each domain (\eg, {\small $K${=}2,3}) achieve superior performance to treating the compound domain as a single target domain (\eg, {\small $K${=}1}). However, performance degradation occurred at unnecessarily large {\small $K$} (\eg, {\small $K${=}5}). Thus, we set {\small $K$} at 3 in all experiments. In addition, we illustrate the effect of continual domain-matching algorithm according to {\small $K$} as \figref{fig:clustering_vis}, which demonstrates our method is capable of adapting to compound domain knowledge.

\paragraph{Distance function~{\small $D_{B}$} for our domain-matching algorithm.} 
To reliably determine pseudo-domain labels with our matching algorithm, selecting the distance function such as \equref{eq:bhata} is crucial. 
So, we conduct a comparative experiment on four types of distance function $D$: Euclidean, Statistics-divergency~\cite{Wang19TransNorm}, Wasserstein~\cite{ruschendorf1985wasserstein}, and Bhattacharyya~\cite{kailath1967divergence}. 
For quantitative comparisons, we calculate domain discrimination ability {\small $dda_l$}, when $l$-layer's statistics of an image defined as {\footnotesize $s_{i}^{l}{=}\{\mu(\phi_{l}(x_i)), \sigma(\phi_{l}(x_i)) \}$}, which is formulated as the following: 
\vspace{-.45em}
\begin{equation}
    \resizebox{.42\hsize}{!}{
    $dda_{l} =\frac{ \mathbb{E}_{d_1 \neq d_2} \:D(s_{m}^{l}, s_{n}^{l}) }{ \mathbb{E}_{d_1 = d_2} \: D(s_{m}^{l}, s_{n}^{l}) },
    $}
  \label{eq:sdr}
\end{equation}
where {\footnotesize $d1, d2 \in \mathbb{D}\,\text{(all domains)}$}, {\footnotesize $m \sim N_{d_1}$}, {\footnotesize $n \sim N_{d_2}$}, and {\footnotesize $N_{d}$} means the number of images of domain {\footnotesize $d$} .
Theoretically, if this rate is high, it means that the function is helpful for distinguishing between different domains.
\figref{fig:dr} shows the domain discrimination ability of each function. The Bhattacharya distance of \equref{eq:bhata} is the most principled choice as it has the highest $dda$ in most cases. In addition, the features from shallow layers are more domain discriminative. Therefore, we utilize the Bhattacharya distance function and set $ddf(x_i)$ as \equref{eq:ddf} for our domain-matching algorithm.

{\renewcommand{\arraystretch}{1.1}
\begin{table}[b]
\vspace{-1.7em}
\caption{\textbf{Experiments with other regularization methods.} Our regularization term produces the best results. `Corr. with mIoU' refers to the correlation between the prediction accuracy and the calculated value by the method.}
\vspace{-1.6em}
\begin{center}
\large{\resizebox{0.35\textwidth}{!}{ 
\begin{tabular}{l x{55}x{55}x{55}}
      & Probability & Entropy & \baseline{Ours}  \\
\shline
mIoU  & 33.1         & 33.3   & \baseline{\textbf{33.7}}  \\
Corr. with mIoU & 0.61         & 0.63   & \baseline{0.68} \\
\end{tabular}
}}\end{center}
\vspace{-1.2em}
\label{tab:regularization}
\end{table}}
{\renewcommand{\arraystretch}{1.1}
\begin{table}[b]
\vspace{-.3em}
\caption{\textbf{Hyperparameter ablation.} We report TTA performance on weather attributes on the C-driving dataset. $\eta$ and $\gamma$ denote the momentum in our matching algorithm and the scale of regularization in \equref{eq:reg_weight}.}
\vspace{-1.9em}
\begin{center}
\large{\resizebox{0.3\textwidth}{!}{ 
\begin{tabular}{x{35} x{35}x{35}x{35}x{35}}

\textbf{{\Large $\eta$}}  & 0.7                  & 0.8                  & 0.9                  & 0.99                  \\
\shline
mIoU & 33.0                 & 33.4                 & \baseline{\textbf{33.7}}                 & 32.3                  \\
\\[-1.em]
\textbf{{\Large $\gamma$}}  & 1                    & 1.5                  & 2                    & 2.5                   \\
\shline
mIoU & 33.5                 & \baseline{\textbf{33.7}}                 & 33.1                 & 32.5                  \\

\end{tabular}
}}\end{center}
\vspace{-.5em}
\label{tab:hyp_ablation}
\end{table}}

\paragraph{Superiority of our regularization term~{\small$\omega(x_i)$}.}
As verified in \tabref{tab:gtav}, our regularization helps the TTA model to control the adaptation pace according to each data in sub-target domains. Moreover, we compare our regularization with several other methods, such as prediction probability and entropy~\cite{entmin}, which can be used to estimate the quality of unsupervised loss from the prediction output. We conduct experiments on whether attributes of the C-driving dataset while keeping our approach but replacing our regularization term with the above methods. \tabref{tab:regularization} demonstrates that our regularization outperforms the competitors and also correlates most strongly with the mIoU score which may represent the quality of unsupervised loss.

\paragraph{Ablation study of the hyperparameters~{\small $\gamma$, $\eta$}.} By experimental analysis in \tabref{tab:hyp_ablation}, we investigate the impact of varying the momentum {\small $\eta$} in our domain-matching algorithm and the scale parameter {\small $\gamma$} in \equref{eq:reg_weight}. We find that setting the momentum to 0.9 produces the best results, whereas a high momentum (e.g., 0.99) may make it challenging to update domain prototypes rapidly, resulting in subpar performance. In addition, we observe that using a large eta strongly minimizes the adaptation loss and hinders adaptation to the target domain, therefore we use the {\small $\eta$} of 1.5.

\section{Conclusion}
In this paper, we propose a simple yet effective TTA approach for dynamic domain changes since previous works suffer from performance degradation in this scenario which is likely to occur in the dynamic world.
We first propose a TTA framework for managing compound domain knowledge via a continual domain-matching algorithm. Our framework helps the TTA model manage multiple domain knowledge by estimating the domain type of an input image among the compound domain. Moreover, in order to prevent overfitting, we modulate the adaptation rates utilizing domain-similarity between the current sub-target and the source domain. With extensive experiments including image classification on ImageNet-C and semantic segmentation on GTA5, C-driving, and Cityscapes, we verify the improved TTA performance of our approach in diverse dynamic TTA scenarios. In this regard, we hope that our efforts will contribute to developing a reliable technique for test-time adaptation and thus facilitating lifelong adaptation in robotics.



\bibliographystyle{plain}
\bibliography{egbib}

\end{document}